\documentclass[sigconf]{acmart}
\usepackage{CJKutf8}
\usepackage[linesnumbered,ruled,vlined]{algorithm2e}
\usepackage{amsmath}

\usepackage{amssymb}
\usepackage{threeparttable}
\usepackage{booktabs,makecell,multirow}

\AtBeginDocument{%
  }

\copyrightyear{2023}
\acmYear{2023}
\setcopyright{rightsretained}
\acmConference[GECCO '23 Companion]{Genetic and Evolutionary Computation Conference Companion}{July 15--19, 2023}{Lisbon, Portugal}
\acmBooktitle{Genetic and Evolutionary Computation Conference Companion (GECCO '23 Companion), July 15--19, 2023, Lisbon, Portugal}\acmDOI{10.1145/3583133.3590618}
\acmISBN{979-8-4007-0120-7/23/07}

\begin{document}
\begin{sloppypar}
\begin{CJK}{UTF8}{min}
\title{A Neural Architecture Search Method using Auxiliary Evaluation Metric based on ResNet Architecture}

\author{Shang Wang}
\affiliation{%
  \institution{Key Laboratory of Intelligent Computing and Information Processing, Ministry of
Education, School of Computer Science \& School of Cyberspace Security, Xiangtan University}
  \city{Xiangtan}
  \state{Hunan}
  \country{China}}
\email{shwang@smail.xtu.edu.cn}

\author{Huanrong Tang}
\authornote{Corresponding author.}
\affiliation{%
  \institution{Key Laboratory of Intelligent Computing and Information Processing, Ministry of
Education, School of Computer Science \& School of Cyberspace Security, Xiangtan University}
  \city{Xiangtan}
  \state{Hunan}
  \country{China}}
\email{tanghuanrong@126.com}
  
\author{Jianquan Ouyang}
\affiliation{%
  \institution{Key Laboratory of Intelligent Computing and Information Processing, Ministry of
Education, School of Computer Science \& School of Cyberspace Security, Xiangtan University}
  \city{Xiangtan}
  \state{Hunan}
  \country{China}}
\email{oyjq@xtu.edu.cn}

\renewcommand{\shortauthors}{Shang Wang, et al.}

\begin{abstract}
This paper proposes a neural architecture search space using ResNet as a framework, with search objectives including parameters for convolution, pooling, fully connected layers, and connectivity of the residual network. In addition to recognition accuracy, this paper uses the loss value on the validation set as a secondary objective for optimization. The experimental results demonstrate that the search space of this paper together with the optimisation approach can find competitive network architectures on the MNIST, Fashion-MNIST and CIFAR100 datasets.
\end{abstract}



\maketitle

\section{Introduction}
Many studies on genetic algorithm-based neural architecture search (GA-NAS) algorithms focus solely on accuracy as the evaluation metric for selecting the best-performing neural network architectures. However, other metrics, such as loss, can also be valuable in identifying potentially good individuals for further training. In this paper, we propose to use the loss value on the validation set as an auxiliary indicator of DNN in the GA-NAS algorithm. Using cooperative optimization in multi-objective genetic algorithms, we search for candidate network structures that perform well on both accuracy and loss metrics. This approach expands the search space and increases the possibility of discovering better neural network architectures.

To support our search, we propose new genetic operators for variable-length gene codes based on the popular ResNet \citep{he2016deep} architecture. After training numerous neural network architectures, those located at the Pareto front \citep{xue2003pareto} undergo additional training to attain the final outcomes.

In our experiments, we compare the multi-objective algorithm \citep{tamaki1996multi} with the single-objective algorithm by adjusting the scale factor of each objective value. This comparison provides insights into the potential benefits of using more than one metric in neural architecture search. Overall, our contributions include improved selection methods, novel genetic operators, and a comprehensive comparison of different optimization algorithms.

\section{Methodology}\label{sec3}
\subsection{Proposed algorithm}\label{subsec2}
Algorithm \ref{algo1} shows the framework of our proposed algorithm. $N$ equals the number of individuals in the population. $T$ commonly manages to be the integer part of $N/5$, which is what we did for our experiments. The function $g_{}^{te}$ in step 10 is the Chebyshev function \citep{zhang2007moea}, which takes the form

\begin{equation}
g^{te}(x\|\lambda,z^*)=\mathop{max}_{1\le i \le m}\{\lambda_i\|f_i(x)-z_i^*\|\} 
\label{1}
\end{equation}

\begin{algorithm}[!h]
\caption{Framework of MO-ResNet}\label{algo1}
\KwIn{stop rule of algorithm; $M$ neural network evaluation metrics; $N$ uniformly distributed weight vectors $\boldsymbol{\lambda_1},\boldsymbol{\lambda_2},...,\boldsymbol{\lambda_N}$; The number of neighbors of each weight vector $T$, the epoch number all individual neural networks trained ${nep}_{train}$, the epoch number EP set individual neural networks trained ${nep}_{full}$.}
\KwOut {set EP.}
$EP\leftarrow \emptyset$\;
for each $i=\{1,2,...,N\}$ let $B_{i}=\{i_1,i_2,...,i_T\}$, where $\boldsymbol{\lambda_{}^{i_1}},\boldsymbol{\lambda_{}^{i_2}},...,\boldsymbol{\lambda_{}^{i_T}}$ are the nearest T vectors to $\boldsymbol{\lambda_{}^{i}}$\;
Initialize N individual ResNet architectures according to the genetic coding strategy and train them to obtain $M$ evaluation indicators,let $\boldsymbol{FV_i}=\boldsymbol{F}(\boldsymbol{x_i})$\;
Initialize $z^*=\{z_1^*,z_2^*,...,z_m^*\}$\;
        \For{$i=1$ to $N$} {
	    Randomly select two indexes k and l from $B_i$, apply crossover and mutation operators to generate new individual $y_i$ from $x_k,x_l$\;
	    Train individual $y_i$ to obtain m evaluation metrics, for each $j=\{1,2,...,m\}$, if $z_j^*<f_j(y_i)$, let $z_j^*=f_j(y_i)$\;
	    Remove all vectors in EP that are dominated by $\boldsymbol{F}(\boldsymbol{y_i})$, and add $\boldsymbol{F}(\boldsymbol{y_i})$ to EP if none of the vectors in EP dominate $\boldsymbol{F}(\boldsymbol{y_i})$\;
        }
        \For{$i=1$ to $N$} {
	    for each $j\in B_i$, if $g_{}^{te}(y_i\| \boldsymbol{\lambda_{}^{j}},z)\le g_{}^{te}(x_j\| \boldsymbol{\lambda_{}^{j}},z)$, let $x_j= y_i$ and $\boldsymbol{FV_j}=\boldsymbol{F}(\boldsymbol{y_i})$\;
        }
If the termination condition is not satisfied, back to line 5\;
Continue to train the individuals in the EP set for ${nep}_{full}$ more epochs.
\end{algorithm}

In Algorithm \ref{algo1}, steps 3 and 7 take the longest time because evaluating
individuals requires training the corresponding neural networks. The algorithm
experiment is based on the BenchENAS platform \citep{xie2022benchenas}, which randomly selects
a GPU to train each individual neural network in the population and summarizes the results in terms of generations. In the original version of the MOEA/D
algorithm \citep{zhang2007moea}, step 11 is placed before step 8, but it is not suitable for this
experiment because the individuals in the population change during the training process, which is not conducive to parallelization. The
implementation details of the proposed algorithm are presented later in this
section.

The training strategy of each individual neural network is shown in Appendix \ref{stra}.

\subsection{Search Space}
Figures of the search space can be found in Appendix \ref{search_space}. Figure \ref{fig:search_space} shows the search space and the individuals encoding strategy. When the crossover operators occur, the convolutional layers and pooling layers in all ResNet blocks are first spliced together (note that the down-sampled $1\times1$ convolutional layer does not participate in the crossover and subsequent mutation operations). Crossover operators occur between convolutional layers, pooling layers or fully connected layers of two individuals. The number of each type of layers may be different, and only the same number of layers will be put crossover operations on. As shown in Figure \ref{fig:crossover_and_mutation} (a)(b)(c), when the number of layers is different, only pairs with short numbers will be applied on crossover operations, and the parts without crossover will be placed in the original position to form offspring individuals.

\subsection{Genetic operators}
In the representation information of neural network individuals, the number of convolutional layers, pooling layers and fully connected layers are encoded as integers, while the detailed information of each layer shown in Figure \ref{fig:search_space} (e.g. $filter\_width$, $output\_channels$) is encoded as real numbers at crossover and mutation, and rounded down when used.

Note that each ResNet block without a skip connection will be applied on the genetic operators in EvoCNN \citep{sun2019evolving}, for it is an one-chain neural network.

To be specific, on a holistic level, genetic operators include: (1) the mutation operation of adding a new layer when the number of its type is lower than the preset upper limit; (2) the mutation operation of randomly removing a layer when the number of its type is higher than the preset lower limit; (3) the mutation operation of changing internal parameters of a layer. When (1) and (2) are triggered, the information representing the number of a certain layer in the encoding changes accordingly. As Figure \ref{fig:crossover_and_mutation} (d)(e)(f) shows, after all mutation operations are performed, if the number of units in an individual is greater than the original one, the additional units form a new ResNet block. Otherwise, units form ResNet blocks in the original way, however, the number of units contained in the last several ResNet blocks may be different. Note that the mutation of FFN Layers does not affect how ResNet blocks are assembled.

While on a detailed level, genetic operators include crossover and mutation operations encoded by each information in Figure \ref{fig:search_space}. The crossover operation uses simulated binary crossover (SBX) \citep{deb1995simulated}, and the mutation operation uses polynomial mutation (PM) \citep{deb2011multi}.

\section{Experiments}
\subsection{Description}
The setup of our experiments is show in Appendix \ref{sec5}. Tables \ref{fig:basic_mnist}-\ref{fig:CIFAR-100} record the performance of the MO-ResNet model on the MNIST, Fashion-MNIST, and CIFAR-100 datasets, respectively. Each best and average error rate was the result of six independent runs of the individual neural network on the EP set. Note that the first dimension of \textbf{fitness\_vector} is the error value and the second dimension is the $k$ times the loss value in Algorithm \ref{algo2}. To show more valid numbers, the fitness\_vector column in Table \ref{fig:basic_mnist} is shown as 1000 times its original value, and the fitness\_vector column in Tables \ref{fig:fashion_mnist} and \ref{fig:CIFAR-100} is shown as 100 times its original value.

In many literatures, NAS is represented as the following bi-level optimization problem:

\begin{equation}
\begin{split}
&\underset{\alpha}{\max}\quad {ACC}_{val}(w^*(\alpha), \alpha),\\
& \begin{array}{r@{\quad}r@{}l@{\quad}l}
s.t.&w^*(\alpha)=\underset{w}{\arg\min}L_{train}(w, \alpha)\\
\end{array} .
\end{split}
\end{equation}

\noindent where ${ACC}_{val}$ is the accuracy on the validation set. The reporting of accuracy varies in different NAS literature, according to our research, some articles report the validation set accuracy and some articles report the test set accuracy. EvoCNN \citep{sun2019evolving} uses the validation set accuracy as the optimization target and calculates the validation set accuracy on the MNIST series dataset in the accompanying code and reports it in the experimental section of the article, while SPOS \citep{guo2020single} explicitly states the experimental results from the test set in the article. For the sake of fairness, the results of the validation set and the test set are reported separately in this paper. During our experiments, we found that these two metrics are similar after sufficient training.

\subsection{Overall results and analysis}
\begin{table*}
\caption{Performance of our model on the MNIST dataset.}\label{fig:basic_mnist}
\begin{tabular}{|l|l|l|l|l|l|l|l|}
\hline
\textbf{model} & \textbf{k} & \textbf{\begin{tabular}[c]{@{}l@{}}|EPset|\end{tabular}} & \textbf{\begin{tabular}[c]{@{}l@{}}fitness\_vector \\ ($\times1000$)\end{tabular}} & \textbf{\#Params} & \textbf{\begin{tabular}[c]{@{}l@{}}best\end{tabular}} & \textbf{\begin{tabular}[c]{@{}l@{}}mean\end{tabular}} &\textbf{\begin{tabular}[c]{@{}l@{}}SE\end{tabular}}\\ \hline
ScatNet-2 \citep{bruna2013invariant} & - & - & - & - & 0.0127 & - & - \\ \hline
BackEISNN \citep{zhao2022backeisnn}  & - & - & - & - & 0.0033 & 0.0042 & 0.0006\\ \hline
EvoCNN \citep{sun2019evolving} & - & - & - & - & 0.0118 & 0.0128 & -\\ \hline
EvoAF \citep{lapid2022evolution} & - & - & - & - & 0.007 & - & -\\ \hline
\multirow{12}{*}{\begin{tabular}[c]{@{}l@{}}MO-ResNet\\ (simplified)\end{tabular}} & 0.0 & 1 & {[}6.5, -{]} & 2.02M & 0.0038/0.0037 & 0.0041/0.0043 & 0.0004/0.0005\\ \cline{2-8} 
 & \multirow{5}{*}{0.2} & \multirow{5}{*}{5} & {[}5.92, 5.83{]} & 1.41M & 0.0041/0.0037 & 0.0043/0.0042 & 0.0003/0.0003\\ \cline{4-8} 
 &  &  & {[}6, 5.55{]} & 1.32M & 0.0039/0.0037 & 0.0041/0.0038 & 0.0002/0.0001\\ \cline{4-8} 
  &  &  & {[}7.08, 5.53{]} & 2.1M & 0.0035/0.0037 & 0.0038/0.0041 & 0.0002/0.0003\\  \cline{4-8} 
   &  &  & {[}5.75, 6.01{]} & 1.27M & 0.0036/0.0040 & 0.0041/0.0043 & 0.0003/0.0002 \\ \cline{4-8} 
    &  &  & {[}5.67, 6.36{]} & 1.63M & 0.0033/0.0038 & 0.0037/0.0042 & 0.0002/0.0003\\ \cline{2-8} 
 & 0.3 & 1 & {[}6, 8.34{]} & 1.46M & 0.0038/0.0039 & 0.0042/0.0041 & 0.0003/0.0002\\ \cline{2-8} 
 & \multirow{5}{*}{0.4} & \multirow{5}{*}{5} & {[}7.5, 12.34{]} & 2.88M & 0.0037/0.0045 & 0.0044/0.0050 & 0.0005/0.0003 \\ \cline{4-8} 
 &  &  & {[}6.83, 12.39{]} & 4.08M & 0.0041/0.0047 & 0.0045/0.0049 & 0.0002/0.0002\\ \cline{4-8} 
  &  &  & {[}6.75, 12.45{]} & 3.69M & 0.0039/0.0043 & 0.0046/0.0048 & 0.0004/0.0003\\ \cline{4-8} 
   &  &  & {[}6.25, 13.01{]} & 2.65M & 0.0041/0.0046 & 0.0044/0.0050 & 0.0003/0.0004\\ \cline{4-8} 
 &  &  & {[}6.42, 12.48{]} & 3.6M & 0.0039/0.0039 & 0.0041/0.0043 & 0.0002/0.0002\\ \hline
\end{tabular}
\end{table*}

\begin{table*}
\caption{Performance of our model on the Fashion-MNIST dataset.}\label{fig:fashion_mnist}
\begin{tabular}{|l|l|l|l|l|l|l|l|}
\hline
\textbf{model} & \textbf{k} & \textbf{\begin{tabular}[c]{@{}l@{}}|EPset|\end{tabular}} & \textbf{\begin{tabular}[c]{@{}l@{}}fitness\_vector\\ ($\times100$)\end{tabular}} & \textbf{\#Params} & \textbf{\begin{tabular}[c]{@{}l@{}}best\end{tabular}} & \textbf{\begin{tabular}[c]{@{}l@{}}mean\end{tabular}} & \textbf{\begin{tabular}[c]{@{}l@{}}SE\end{tabular}} \\ \hline
GoogleNet \citep{szegedy2015going} & - & - & - & 5.98M & 0.0786 & - & -\\ \hline
BackEISNN-E \citep{zhao2022backeisnn}  & - & - & - & - & 0.073 & 0.0744 & 0.0008\\ \hline
BackEISNN-D \citep{zhao2022backeisnn}  & - & - & - & - & 0.0655 & 0.0696 & 0.0031\\ \hline
MCNN15 \citep{nocentini2022image} & - & - & - & - & 0.0596 & - & -\\ \hline
EvoCNN \citep{sun2019evolving} & - & - & - & 6.52M & 0.0547 & 0.0728 & -\\ \hline
EvoAF \citep{lapid2022evolution} & - & - & - & - & 0.1025 & - & -\\ \hline
\multirow{4}{*}{MO-ResNet} & 0.0 & 1 & {[}6.37, -{]} & 3.52M & 0.0409/0.0440 & 0.0424/0.0449 & 0.0009/0.0009\\ \cline{2-8} 
 & 0.2 & 1 & {[}6.85, 3.8{]} & 5.45M & 0.0448/0.0468 & 0.0455/0.0479 & 0.0007/0.0009\\ \cline{2-8} 
 & 0.3 & 1 & {[}6.14, 5.11{]} & 50.97M & 0.0424/0.0463 & 0.0438/0.0469 & 0.0009/0.0006\\ \cline{2-8} 
 & 0.4 & 1 & {[}6.28, 7.27{]} & 1.96M & 0.0414/0.0442 & 0.0424/0.0457 & 0.0006/0.0012
 \\ \hline
\end{tabular}
\end{table*}

\begin{table*}
\caption{Performance of our model on the CIFAR-100 dataset.}\label{fig:CIFAR-100}
\begin{tabular}{|l|l|l|l|l|l|l|l|}
\hline
\textbf{model} & \textbf{k} & \textbf{\begin{tabular}[c]{@{}l@{}}|EPset|\end{tabular}} & \textbf{\begin{tabular}[c]{@{}l@{}}fitness\_vector\\ ($\times100$)\end{tabular}} & \textbf{\#Params} & \textbf{\begin{tabular}[c]{@{}l@{}}best\end{tabular}} & \textbf{\begin{tabular}[c]{@{}l@{}}mean\end{tabular}} & \textbf{\begin{tabular}[c]{@{}l@{}}SE\end{tabular}}\\ \hline
ResNet-18 \citep{he2016deep} & - & - & - & 11.22M & 0.2575 & - & -\\ \hline
ResNet-50 \citep{he2016deep} & - & - & - & 23.71M & 0.2709 & - & -\\ \hline
PRE-NAS \citep{peng2022pre} & - & - & - & - & 0.2651/0.2649 & 0.2805/0.2798 & 0.0121/0.0122\\ \hline
\multirow{12}{*}{\begin{tabular}[c]{@{}l@{}}MO-ResNet \end{tabular}} & 0.0 & 1 & {[}32.3, -{]} & 63.41M & 0.2362/0.2414 & 0.2406/0.2440 & 0.0028/0.0044\\ \cline{2-8} 
 & \multirow{4}{*}{0.2} & \multirow{4}{*}{4} & {[}32.54, 26.26{]} & 55.74M & 0.2460/0.2486 & 0.2488/0.2517 & 0.0027/0.0023\\ \cline{4-8} 
 &  &  & {[}32.8, 25.82{]} & 6.72M & 0.2395/0.2414 & 0.2455/0.2490 & 0.0035/0.0036\\ \cline{4-8} 
  &  &  & {[}32.49, 26.29{]} & 7.01M & 0.2426/0.2456 & 0.2472/0.2477 & 0.0029/0.0024\\  \cline{4-8} 
   &  &  & {[}32.77, 26.23{]} & 56.13M & 0.2482/0.2496 & 0.2494/0.2522 & 0.0009/0.0018\\ \cline{2-8} 
 & \multirow{5}{*}{0.3} & \multirow{5}{*}{5} & {[}34.13, 38.51{]} & 17.43M & 0.2577/0.2579 & 0.2615/0.2619 & 0.0025/0.0039\\ \cline{4-8} 
    &  &  & {[}32.89, 39.04{]} & 15.88M & 0.2568/0.2603 & 0.2613/0.2622 & 0.0025/0.0021\\ \cline{4-8} 
   &  &  & {[}33.6, 38.53{]} & 15.55M & 0.2564/0.2579 & 0.2588/0.2626 & 0.0019/0.0033\\ \cline{4-8} 
 &  &  & {[}33.38, 38.62{]} & 14.81M & 0.2590/0.2614 & 0.2631/0.2640 & 0.0027/0.0021\\ \cline{4-8} 
  &  &  & {[}33.03, 38.63{]} & 13.06M & 0.2561/0.2559 & 0.2582/0.2603 & 0.0019/0.0028\\ \cline{2-8} 
    & \multirow{2}{*}{0.4} & \multirow{2}{*}{2} & {[}33.2, 51.45{]} & 66.88M & 0.2425/0.2492 & 0.2463/0.2531 & 0.0025/0.0025\\ \cline{4-8} 
 &  &  & {[}32.09, 51.66{]} & 67.47M & 0.2435/0.2484 & 0.2464/0.2498 & 0.0031/0.0015\\ \hline
\end{tabular}
\end{table*}

Experimental outcomes were acquired by configuring various $k$ values. Tables \ref{fig:basic_mnist}-\ref{fig:CIFAR-100} display the best error rate and the average error rate in columns 6-7. Data before the slanted line indicates validation set results, and data after refers to test set values. It is worth mentioning that when $k=0$, the algorithm solely explored the image recognition accuracy metric.

As demonstrated in the tables, MO-ResNet outperformed Scat-Net \citep{bruna2013invariant} in terms of recognition accuracy on the MNIST dataset when compared to a hand-designed neural network. However, it didn't improve upon BackEISNN \citep{zhao2022backeisnn} concerning the best results, yet it surpassed it in average error rates acquiring when adjusting certain $k$ values. The algorithm proposed in this paper attained significantly superior model recognition accuracy compared to the GoogleNet \citep{szegedy2015going}, BackEISNN \citep{zhao2022backeisnn}, and MCNN15 \citep{nocentini2022image} models on the Fashion-MNIST dataset. Moreover, MO-ResNet outperformed the two fundamental ResNet \citep{he2016deep} models on the CIFAR100 dataset. Compared to other NAS techniques, MO-ResNet had better performance than EvoCNN \citep{sun2019evolving} and EvoAF \citep{lapid2022evolution} on the MNIST and Fashion-MNIST datasets, and outperformed PRE-NAS \citep{peng2022pre} on the CIFAR100 dataset.

Moreover, it is noteworthy that the best error rate for the MNIST dataset was not achieved when $k=0$. Furthermore, models with fewer parameters and a comparable or better accuracy than other models could be obtained on both the Fashion-MNIST and CIFAR100 datasets when $k$ was not zero. These observations suggest that auxiliary evaluation metrics could increase the probability of discovering a competitive network that balances the trade-off between parameters and accuracy.

Appendix \ref{runtime} shows the runtime analysis.

\subsection{Transfer learning}
We migrated the optimal network architecture searched on the CIFAR-100 dataset starting with $7\times 7$ convolution instead of $3\times 3$, and then ran 150,000 steps on the imagenet dataset, obtaining 39.8$\%$ Top-1 error and 18.5$\%$ Top-5 error on the training set, and on the The test set yielded a Top-1  error of 35.4$\%$ and a Top-5 error of 14.2$\%$, for a total of approximately 36 hours of training on 8 a100 GPU graphics cards. It may be possible to try to migrate more of the network structures obtained from training on the CIFAR-100 dataset in the future work.

\section{conclusion}
In this paper, we propose a neural architecture search algorithm MO-ResNet based on the ResNet architecture, which is based on the MOEA/D algorithm. After the search is completed, the individual neural networks on the EP set are further trained and the results are compared with the hand-designed neural networks and some other NAS search algorithms. Competitive results were achieved on the MNIST, Fashion-MNIST, and CIFAR-100 datasets. Future work could consider more neural network evaluation metrics and design more combinations of objective functions to find better neural network architectures. \footnote{ The code of this paper is published on https://github.com/ra225/code\_on\_BenchENAS.}

\begin{acks}
This research has been supported by Key Projects
of the Ministry of Science and Technology of the People Republic of China
(No.2020YFC0832405).
\end{acks}

\bibliographystyle{ACM-Reference-Format}
\bibliography{sample-base}

\newpage
\newpage
\begin{appendix}
\section{Related work}
EvoCNN algorithm \citep{sun2019evolving} first proposed a variable length encoding strategy in the development of GA-NAS algorithms. The search space of EvoCNN is an one-chain neural network structure, and in the experimental part of \citep{sun2019evolving}, the test result on MNIST series datasets of EvoCNN is reported. 

The search space of our work is an extension of EvoCNN. As Figure \ref{fig:search_space} shows, without shortcut connections, the ResNet block will degenerate to an one-chain structure, so we could apply EvoCNN's genetic operator to each resnet block, then add a shortcut connection to it, and $1\times1$ convolutional layers for downsampling in the shortcut connection will depend on the case.

In the experimental section, EvoCNN performed a single-objective genetic algorithm for NAS on 9 datasets, 6 of which were MNIST series. Our algorithm performed a multi-objective NAS on 3 datasets, including ablation experiments, with 4 adjustments of the target coefficients on each dataset, thus corresponding to a single-objective search on 12 datasets. Since our search space is based on ResNet, experiments were performed on CIFAR-100, a three-channel dataset, in addition to 2 datasets of the MNIST series for comparison with EvoCNN.

Our experiments are based on the BenchENAS platform \citep{xie2022benchenas}, another work of the EvoCNN authors, to whom we would like to express our gratitude.

\section{Training strategy}
\label{stra}
Algorithm \ref{algo2} describes the procedure for training individual neural networks and calculating the best fitness values. By setting $batch\_size$, the loop body from step 2 to step 19 trains each individual neural network in the population and obtains the fitness vector.
\begin{algorithm}[!h]
	\caption{Multi-objective fitness calculation}\label{algo2}
	\KwIn{Population $P_t$; training epoch number $T$; the training set $D_{train}$; the validation set $D_{val}$; the size of a mini-batch $batch\_size$; learning rate $lr$; the fitness coefficient $k$; classifier}
	\KwOut{the fitness of the population $P_t$.}
	\For {each s in $P_t$} {
    	$best\_error\leftarrow 1$\;
    	$best\_loss\leftarrow +\infty$\;
    	$train\_step\leftarrow \|D_{train}\|/batch\_size$\;
    	$eval\_step\leftarrow \|D_{val}\|/batch\_size$\;
    	Decode individual $s$ into the original ResNet structure, let $\boldsymbol{W}=\{w_1,w_2,...\}$ be the weight set of the network, initialize $\boldsymbol{W}$\;
    	\For{$i=1$ to $T$} {
    	    \For{$j=1$ to $train\_steps$} {
    	      $\boldsymbol{W}=\boldsymbol{W}-lr\times \nabla_{\boldsymbol{W}}L(\boldsymbol{W},D_{train,j})$\;
    	        \tcc{$L(\boldsymbol{W},D_{train,j})$ is the classifier's loss value of the j-th batch data $D_{trainj}$ under $\boldsymbol{W}$}
    	    }
    	    $sum\_loss\leftarrow 0$\;
    	    $correct\leftarrow 0$\;
    	    \For{$j=1$ to $eval\_steps$} {
    	        $sum\_loss\leftarrow sum\_loss+L(\boldsymbol{W},D_{val,j})$\;
    	        $correct\leftarrow correct+{acc}_j$\;
    	        \tcc{${acc}_j$ is the classifier's prediction accuracy of the j-th batch data $D_{val,j}$ under $\boldsymbol{W}$}
    	    }
    	    \If {$1-correct/\|D_{val}\|<best\_error$} {
    	        $best\_error=1-correct/\|D_{val}\|$\;
    	    }
    	    \If {$sum\_loss/\|D_{val}\|<best\_loss$} {
    	        $best\_loss=sum\_loss/\|D_{val}\|$\;
    	    }
    	}
        $\boldsymbol{s.fitness}\leftarrow (best\_error, k\times best\_loss)$
    }
\end{algorithm}

In algorithm \ref{algo2}, stochastic gradient descent is used to train the neural network, and the direction of the gradient is theoretically the fastest decreasing of the classification error value. Due to the learning rate, the error value is generally dominated by the direction of descent during the training process, but there are often cases when the error value rises instead of falling after one step of updating the neural network weights.

To accurately evaluate each individual neural network, in addition to training a specific number of epochs, the current loss value and accuracy will be calculated after each epoch training and the historical optimal value of the individual neural network will be updated. Step 3, step 7 and step 12 of Algorithm \ref{algo1} all call Algorithm \ref{algo2}, where steps 3 and 7 pass $T={nep}_{train}$ and step 12 passes $T={nep}_{full}$.

\section{Figure of search space}
\label{search_space}
\begin{figure*}
\centering
\includegraphics[scale=0.65]{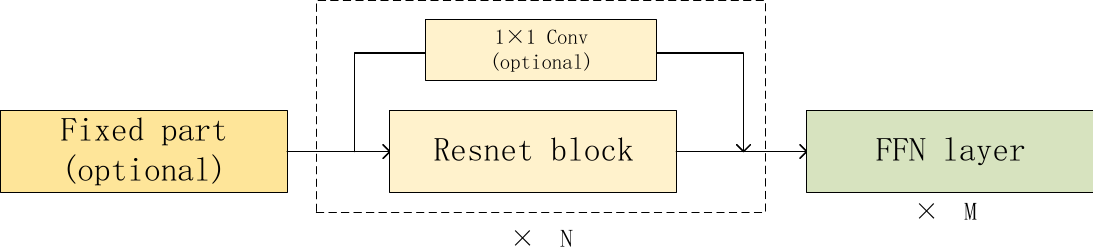}
\begin{center}
\captionsetup{singlelinecheck=off}
\caption{
Illustration of search space. The fixed part and the ResNet block consist of convolutional layers and pooling layers. In the process of initialization, the number of ResNet blocks N and the number of FFN layers M are randomly generated within a preset range. Then, the number of convolutional and pooling layers in each ResNet block is also randomly generated within another preset range. 
Last but not least, the hyperparameters of each conv/pool/FFN layer are also randomly selected within a predefined range.
Since for every individual, the fixed part will be the same, it does not need to be encoded. The encoding strategy of the individual could be recursively defined using pseudo regular expressions as follows
\begin{align*}
    {\bf gene\_encode} &\rightarrow {\bf resnet\_block\_encode}^N\ {\bf FFN\_layer\_encode}^M \\
    {\bf resnet\_block\_encode} &\rightarrow ({\bf conv\_layer\_encode} \ \| \ {\bf pool\_layer\_encode})^+ \\
    {\bf conv\_layer\_encode} &\rightarrow [{\bf conv\_identify},\ (filter\_width,\  filter\_height), (stride\_width,\ stride\_height),\ output\_channels] \\
    {\bf pool\_layer\_encode} &\rightarrow [{\bf pool\_identify},\ (filter\_width,\  filter\_height), (stride\_width,\ stride\_height),\ pool\_type] \\
    {\bf FFN\_layer\_encode} &\rightarrow [{\bf FFN\_identify},\ output\_neurons] \\ 
    {\bf conv\_identify} &\rightarrow 1,\ {\bf pool\_identify} \rightarrow 2,\ {\bf FFN\_identify} \rightarrow 3 
\end{align*}
The encoding position $pool\_type$ indicates maximum or average pooling. Note that in a ResNet block, when the input and the output have different lengths, widths, or number of channels, a $1\times1$ conv is required to downsample the input.
}
\label{fig:search_space}
\end{center}
\end{figure*}

\begin{figure*}
\centering
\includegraphics[scale=0.51]{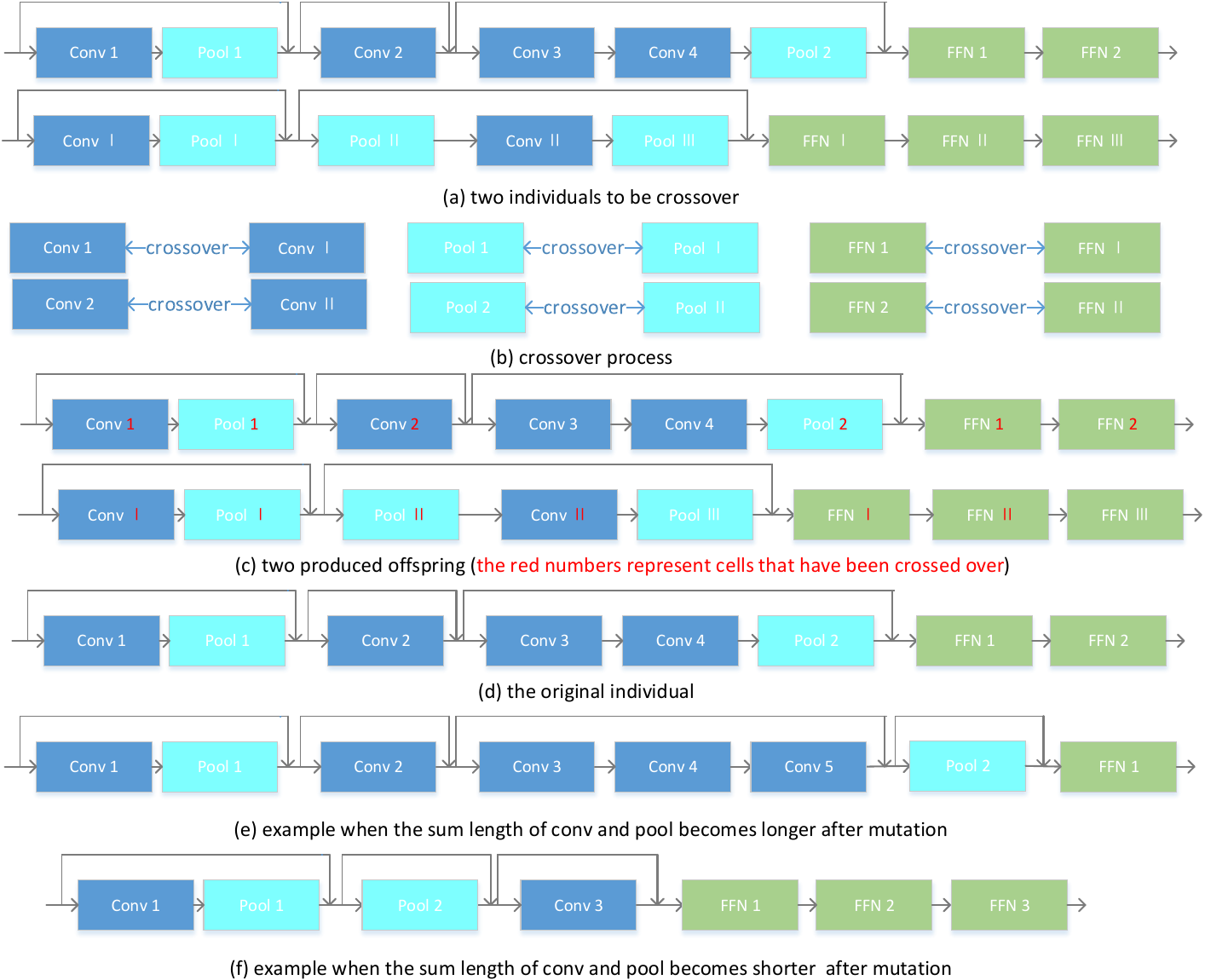} \caption{Illustration of genetic operators between individuals. (a)(b)(c) show the crossover, while (d)(e)(f) show the mutation.}
\label{fig:crossover_and_mutation}
\end{figure*}

See Figure \ref{fig:search_space} and Figure \ref{fig:crossover_and_mutation}.

\section{Experimental setup}\label{sec5}
\subsection{Environment configuration}
The experiment in this paper is based on BenchENAS platform \citep{xie2022benchenas}, which can be deployed in a multi-machine and multi-card GPU environment with one central node server and multiple working nodes.
\begin{figure}
\centering
\includegraphics[scale=0.5]{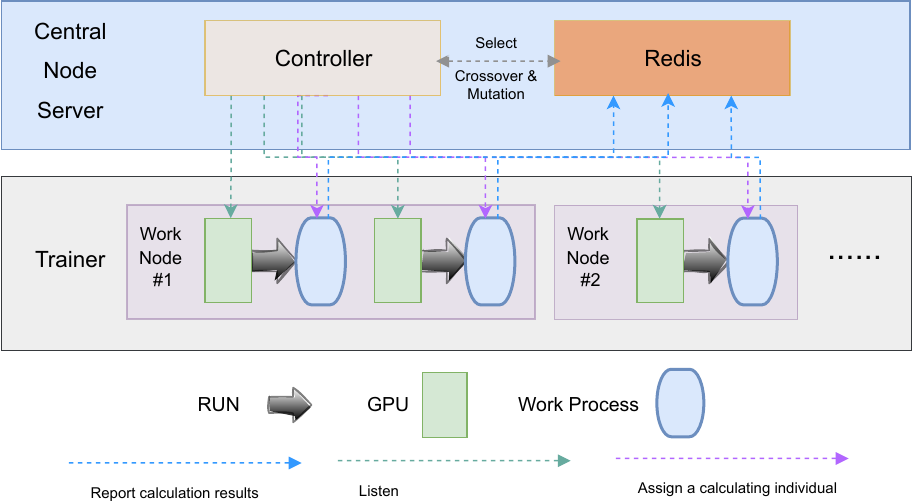} \caption{A brief illustration of BenchENAS platform deployment. The central node server is composed of the controller part and redis. After the individuals of each population are generated, the controller distributes all individual neural networks to the worker nodes. In the training process, the controller monitors the GPU status of the working node. When a GPU with sufficient video memory and suitable for training appears, a work node is ordered to create a process on the GPU to train an individual neural network. After the training is completed, the work node reports the result to the redis of the central node. After receiving the redis write event, the central node records the reported data in a disk file. According to the training results, the central node makes selection, crossover and mutation operations. When the training process of an individual ends, the GPU memory occupied will be released and allocated to the next individual. The central node server does not perform specific training tasks.}
\label{fig:BenchENAS}
\end{figure}

In this paper, two working nodes are used, graphics card type is NVIDIA GeForce RTX 3090, the programming language is Python 3.6.8, the deep learning framework is Pytorch \citep{paszke2019pytorch} 1.10.1. The operating system of central node server is Ubuntu 18.04, and that of work nodes are all Ubuntu 20.04.
\subsection{Settings}
The Environment configuration is shown in .The MNIST \citep{lecun1998mnist},   Fashion-MNIST \citep{xiao2017fashion} and CIFAR-100 \citep{krizhevsky2019cifar} datasets are used to test the effect of neural architecture search. For all datasets, the crossover probability is set 0.9, all mutation operators probability is set 0.2.   In algorithm \ref{algo1}, $M$ is set to 2, and $M$ neural network evaluation metrics are the accuracy and loss value of the validation set, respectively. The optimizer used is SGD with $momentum=0.9$ and $weight\_decay=0.0005$.

The search space of the MNIST dataset contains only one ResNet block without shortcut connection plus fully connected layers, i.e., a one-chain neural network structure, because the MNIST dataset is easier to identify and the search space is simpler. For other datasets, the number of ResNet blocks is chosen randomly between [1,10]. The fixed part, a 64-channel $3\times3$ convolution followed by a BatchNorm \citep{ioffe2015batch} layer and a Relu \citep{glorot2011deep} layer, is included only for the Fashion-MNIST and CIFAR-100 datasets. In fact, all convolutional layers in the ResNet block are also followed by a BatchNorm layer and a Relu layer.

\begin{figure}
\centering
\includegraphics[scale=0.6]{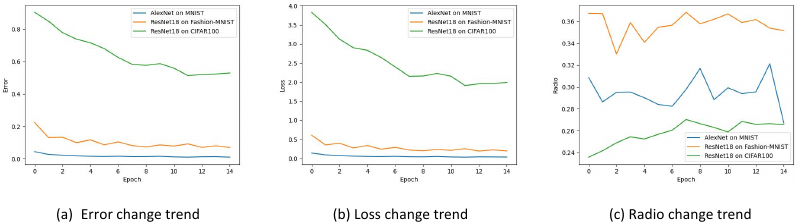} \caption{Figure of the change trend of error, loss and ratio (i.e. $error/loss$) values within 15 epochs. }
\label{fig:show_ratio}
\end{figure}

It can be seen from Figure \ref{fig:show_ratio}(c) that the value of ratio is in the interval $[0.2,0.4]$ on all datasets, so in comparison experiments, $k$ in in algorithm \ref{algo2} is set to 0.2, 0.3 and 0.4 respectively. 

In the MNIST dataset, both the number of generations and the number of individuals were set to 50, which is equal to EvoCNN \citep{sun2019evolving}. i.e. for each value of $k$, 2500 neural network individuals will be evaluated. Due to the longer training time and larger memory occupation of neural networks with shortcut connections, both counts were set to 20 in the Fashion-MNIST dataset and 35 in the CIFAR100 dataset. i.e., for each value of $k$, 400 individuals were trained on the Fashion-MNIST dataset and 1225 individuals on the CIFAR100 dataset, but this did not prevent us from achieving a competitive experimental result. In the experimental part of EvoCNN, the number of generations was set to 50 for all datasets including Fashion-MNIST, and we found individuals with better performance than them on the Fashion-MNIST dataset by evaluating less than one-sixth of this number of individuals with residual connections.

Some other settings are shown in Table \ref{tab1}. During the training of all individuals, i.e. steps 3 and 7 of Algorithm \ref{algo1}, $lr$ is updated using CosineAnnealingLR \citep{loshchilov2016sgdr} after initialization. During the retraining of elite individuals, i.e. step 12 of Algorithm \ref{algo1}, after $lr$ is re-initialised, the minimum loss value during the current training is recorded and if no smaller loss value appears after 8 epochs of training, $lr$ is halved.
\begin{table}[h]
\begin{center}
\caption{Some settings for each dataset.}\label{tab1}%
\begin{tabular}{@{}llll@{}}
\toprule
\multirowcell{2}{Parameter} & \multirowcell{2}{MNIST} & Fashion- & \multirowcell{2}{CIFAR-100}\\
                            &    &MNIST &  \\
\midrule
$batch\_size$ in algorithm \ref{algo2}      & 64   &  100 & 100  \\
\midrule
initial $lr$     & 0.025   & 0.025  & 0.1  \\
\midrule
${nep}_{train}$     & 10   & 12  & 45  \\
\midrule
${nep}_{full}$     & 90   & 100  & 260  \\
\midrule
\# of conv layers     & select from                     & select from    \\ 
\quad per ResNet block &  \quad [1,4] & \quad  [1,3]\\
\midrule
\# of pool layers     & select from & \\
\quad  in one full individual & \quad [1,3] & \\
\midrule
\# of FFN layers  & select from & select from \\
\quad  in one full individual & \quad [1,4] &\quad [1,2] \\
\midrule
range of filter\_size  & \multirowcell{2}{[2,20]} & \multirowcell{2}{[1,5]} \\
\quad  for conv layers & & \\
\midrule
range of output\_neurons  & \multirowcell{2}{[1000,2000]}\\
 \quad for FFN layers & \\
 \midrule
available filter\_size set & \multirowcell{2}{\{2,4\}}\\
\quad for pool layers & \\
\midrule
range of \#channels  & \multirowcell{2}{[3,50]} & \multirowcell{2}{[3,128]} \\
\quad for conv layers & \\
\bottomrule
\end{tabular}
\end{center}
\end{table}

\section{Runtime analysis}
\label{runtime}
The algorithm needs to train $nep\_train$ epochs for each $k$ value for each individual neural network in each generation. Assuming a neural network is trained on a dataset for 1 epoch as a time unit, a total of $50\times 50\times 10\times 4=100000$ time units are run on the MNIST dataset, $20\times 20\times 12\times 4=19200$ time units are run on the Fashion-MNIST dataset and $35\times 35\times 45\times 4=220500$ time units are run on the CIFAR100 dataset. However, since we did not introduce residual connectivity on the MNIST dataset, the actual time required to train an individual neural network on the MNIST dataset for 1 time unit was shorter.

In practice, one graphics card can train multiple individual neural networks at the same time, and the NAS training process for multiple datasets can be carried out simultaneously without taking up full graphics memory. The use of multi-machine multi-card distributed training is often affected by network communication fluctuations, changes in the number of graphics card resources, host performance and other uncertainties, if GPU time is calculated based on the number of $graphics cards\times run time$, the total number of GPU days run for the MNIST dataset experiment is about 6\textasciitilde 7, the total number of GPU days run for the Fashion-MNIST dataset experiment is about 4, and the total number of GPU days run for the CIFAR100 dataset experiment is about 37\textasciitilde 39.

\end{appendix}


\end{CJK}
\end{sloppypar}
\end{document}